# Random crossings in dependency trees

*Ramon Ferrer-i-Cancho*[1]

**Abstract.** It has been hypothesized that the rather small number of crossings in real syntactic dependency trees is a side-effect of pressure for dependency length minimization. Here we answer a related important research question: what would be the expected number of crossings if the natural order of a sentence was lost and replaced by a random ordering? We show that this number depends only on the number of vertices of the dependency tree (the sentence length) and the second moment about zero of vertex degrees. The expected number of crossings is minimum for a star tree (crossings are impossible) and maximum for a linear tree (the number of crossings is of the order of the square of the sequence length).

*Keywords: syntactic dependency trees, syntax, distance, crossings, planarity.*

## 1. INTRODUCTION

According to dependency grammar (Mel'čuk 1988, Hudson 2007) the structure of a sentence can be defined by means of a tree in which vertices are words and arcs indicate syntactic dependencies between these words (Fig. 1). Here we focus on the crossings between dependencies due to the linear arrangement of the vertices of a tree (Hays 1964, Holan et al. 2000, Hudson 2000, Havelka 2007).

Imagine that $\pi(v)$ is the position of vertex $v$ in linear arrangement of the vertices of a tree, a number between 1 and $n$, with $n$ being the length of the sequence. Imagine that we have two pairs of linked vertices: $(u,v)$ and $(s,t)$, such that $\pi(u) < \pi(v)$ and $\pi(s) < \pi(t)$. The arcs (or edges) defined respectively by $(u,v)$ and $(s,t)$ cross if and only if

$$\pi(u) < \pi(s) < \pi(v) < \pi(t) \qquad (1)$$

or

$$\pi(s) < \pi(u) < \pi(t) < \pi(v). \qquad (2)$$

$C$ is defined as the number of different pairs of edges that cross. For instance, $C = 0$ in the sentence in Fig. 1 and $C = 9$ in Fig. 2. When there are no vertex crossings ($C = 0$), the syntactic dependency tree of a sentence is said to be planar (Havelka 2007).

According to crossing theory, $C$ cannot exceed $C_{pairs}$, the number of edge pairs that can potentially cross, which is (Ferrer-i-Cancho 2013)

$$C_{pairs} = \frac{n}{2}\left(n - 1 - \langle k^2 \rangle\right), \qquad (3)$$

---

[1] Complexity and Quantitative Linguistics Lab. Departament de Ciències de la Computació, LARCA Research Group, Universitat Politècnica de Catalunya (UPC). Campus Nord, Edifici Omega, Jordi Girona Salgado 1-3, 08034 Barcelona, Catalonia (Spain). Phone: +34 934134028. Fax: +34 934137787. E-mail: rferrericancho@cs.upc.edu

where $n$ is the sequence length (the number of words/vertices) and $\langle k^2 \rangle$ is the second moment about zero of the degree, defined as

$$\langle k^2 \rangle = \frac{1}{n} \sum_{i=1}^{n} k_i^2, \qquad (4)$$

where $k_i$ is the degree of the $i$-th vertex of the tree. As the first moment of the degree of a tree of $n$ vertices is constant, i.e. $\langle k \rangle = 2 - 2/n$ (Noy 1998), the degree variance of a tree is fully determined by $\langle k^2 \rangle$ and $n$.

For the dependency tree of Fig. 1, Eq. 3 gives $C_{pairs} = 18$ since $n = 9$ and $\langle k^2 \rangle = 4$.

It has been argued that the small amount of crossings in real sentences (Liu 2010) could be a side-effect of a principle of dependency length minimization (Ferrer-i-Cancho 2006, Ferrer-i-Cancho 2013). A challenge for this hypothesis is that the number of crossings that is expected by chance (by ordering the vertices at random) is about the same value that is obtained in real sentences. Thus, a theoretical analysis of E[$C$], the expected number of crossings in a random linear arrangement of vertices is needed to shed light on the statistical significance of the rather low number of crossings in real sentences (Liu 2010). This is the goal of the next sections: Section 2 reviews previous results on the maximum value of $C$ and Section 3 derives E[$C$] = $C_{pairs}/3$, and related results, e.g., the probability that two edges cross when arranged linearly at random. If the edges share no vertex the probability is 1/3 and it is zero otherwise. Section 4 discusses some applications of these results.

## 2. CROSSING THEORY

$u \sim v$ is used to refer to the edge defined by the pair of vertices $(u,v)$. The edges $u \sim v$ and $s \sim t$, such that $u < v$ and $s < t$, cannot cross if they have a vertex in common, i.e. $u \in \{s,t\}$ or $v \in \{s,t\}$. Therefore $C > 0$ requires that there is at least a pair of edges that are formed by four different vertices. Thus $C = 0$ if $n < 4$ and $C > 0$ needs $n \geq 4$.

The structure of a tree, e.g., a syntactic dependency tree, can be defined by means of an adjacency matrix $A = \{a_{uv}\}$, where $a_{uv} = 1$ if the pair of vertices $(u,v)$ is linked and otherwise $a_{uv} = 0$. The matrix is symmetric $a_{uv} = a_{vu}$ (the direction of a dependency is neglected). Loops are not allowed ($a_{uu} = 0$). $a_{uv} = 1$ and $u \sim v$ are equivalent.

The number of crossings induced by the linear arrangement of the vertices can be defined as

$$C = \frac{1}{4} \sum_{u=1}^{n} \sum_{v=1}^{n} a_{uv} C(u,v), \qquad (5)$$

where $C(u,v)$ is the number of different edges that cross with the edge $u \sim v$. By symmetry, $C(u,v) = C(v,u)$. The factor 1/4 of Eq. 5 comes from the fact that the same crossing is counted four times in that formula:
- Two times due to the double summation of Eq. 5, i.e. the target edge $u \sim v$ is counted first through the pair $(u, v)$ and second through its symmetric pair $(v, u)$.
- Two times more due to the fact the edges of the form $s \sim t$ with which the edge $u \sim v$ crosses are counted twice, first through $C(u,v)$ and second through $C(s,t)$.

$C(u,v)$ can be defined in turn as

$$C(u,v) = \frac{1}{2} \sum_{\substack{s=1 \\ s \neq u,v}}^{n} \sum_{\substack{t=1 \\ t \neq u,v}}^{n} a_{st} C(u,v;s,t), \qquad (6)$$

where $C(u,v;s,t) = 1$ if the edge $u\sim v$ crosses the edge $s\sim t$ and $C(u,v;s,t) = 0$ otherwise. The factor 1/2 in Eq. 6 comes from the fact that an edge is encountered twice in the double summation, first by the pair of vertices $(s,t)$ and second by the pair $(t, s)$.

It has been argued that $C(u,v)$ cannot exceed $C_{pairs}(u,v) = n - k_u - k_v$ where $k_x$ is the degree of vertex $x$ (Ferrer-i-Cancho 2013; see Appendix A of the present article for a derivation of $C_{pairs}(u,v)$). Thus the total number of crossings of the linear arrangement of a tree cannot exceed (Ferrer-i-Cancho 2013)

$$C_{pairs} = \frac{1}{4} \sum_{u=1}^{n} \sum_{v=1}^{n} a_{uv} C_{pairs}(u,v) = \frac{n}{2}\left(n - 1 - \left\langle k^2 \right\rangle\right). \qquad (7)$$

A star tree is a tree with a vertex of maximum degree while a linear tree is a tree where the maximum vertex degree is two (Fig. 3). Linear and star trees are important trees for crossing theory as they determine the range of variation of $\left\langle k^2 \right\rangle$ in Eq. 7. $\left\langle k^2 \right\rangle$ is minimized by a linear tree (Ferrer-i-Cancho 2013) and that tree is indeed the only minimum (Appendix B). Similarly, $\left\langle k^2 \right\rangle$ is maximized by a star tree (Ferrer-i-Cancho 2013) and that tree is indeed the only maximum (Appendix B).

A very simple case to demonstrate Eq. 7 is a linear tree with $n = 4$. That tree has three edges and two leaves (a leaf is a vertex of degree one). Imagine that the two leaves are labeled with 1 and 4 and the other edges are labeled with 2 and 3. The only pair of edges that can cross are 1~2 and 3~4 (the two different edges formed by each of the two leaves), since they are the only pair of edges that do not share vertices. Thus $C_{pairs} = 1$ and $C$ is binary, i.e. $C = 1$ (edges 1~2 and 3~4 cross) or $C = 0$ (edges 1~2 and 3~4 do not cross). Accordingly, applying $n = 4$ and $\left\langle k^2 \right\rangle = (1+1+4+4)/4 = 5/2$ to Eq. 7 yields $C_{pairs} = 1$ for that linear tree.

## 3. RANDOM CROSSINGS

According to Eq. 5, the expected number of crossings induced by a random linear arrangement of the vertices is

$$E[C] = \frac{1}{4} \sum_{u=1}^{n} \sum_{v=1}^{n} a_{uv} E[C(u,v)] \qquad (8)$$

while the expectation of $C(u,v)$ is in turn

$$E[C(u,v)] = \frac{1}{2} \sum_{\substack{s=1 \\ s \neq u,v}}^{n} \sum_{\substack{t=1 \\ t \neq u,v}}^{n} a_{st} E[C(u,v;s,t)]. \qquad (9)$$

As $C(u,v;s,t)$ is an indicator variable, $E[C(u,v;s,t)] = p_c(u,v;s,t)$, the probability that the edges $u\sim v$ and $s\sim t$ cross knowing that $s \notin \{u,v\}$ and $t \notin \{u,v\}$. By the definition of crossing in Eqs. 1 and 2, it follows that $p_c(u,v;s,t) = 0$ if the edges $u\sim v$ and $s\sim t$ have at least one vertex in common, i.e. $u \in \{s,t\}$ or $v \in \{s,t\}$. Otherwise, $p_c(u,v;s,t) = 1/3$. To see the latter, notice that the random linear arrangement of two edges is equivalent to:

- Generating four different vertex positions with the only constraint that they are random numbers between 1 and $n$ and positions that are not taken yet are equally likely.
- Sorting the four positions increasingly giving $\pi_1$, $\pi_2$, $\pi_3$ and $\pi_4$ such that $1 \leq \pi_1 < \pi_2 < \pi_3 < \pi_4 \leq n$. It is said that $\pi_i$ has rank $i$.
- Assigning each of these four positions to a different vertex of the pairs of edges involved. Eqs. 1 and 2 mean that the two edges cross if and only if $(u,v)$ is assigned $(\pi_1, \pi_3)$ or $(\pi_2, \pi_4)$.

Therefore the probability that $u\sim v$ and $s\sim t$ cross is the probability of assigning two of the four positions whose ranks are not consecutive to the vertices of $u\sim v$ with $u < v$, i.e. (a) $\pi(u) = \pi_1$ and $\pi(v) = \pi_3$ or (b) $\pi(u) = \pi_2$ and $\pi(v) = \pi_4$. Therefore,

$$p_c(u,v;s,t) = \frac{2}{\binom{4}{2}} = \frac{1}{3}. \qquad (10)$$

Interestingly, the probability that two edges cross does not depend on the sequence length $n$ once it is known whether they share vertices or not (if the two edges share vertices the probability is zero regardless of $n$; if they do not share any vertex then $n \geq 4$ and the probability is 1/3). Furthermore, the identity of vertices involved is irrelevant for the probability that they cross once it is known if the edges share vertices or not. Thus, Eq. 9 becomes

$$E[C(u,v)] = \frac{C_{pairs}(u,v)}{3}. \qquad (11)$$

Applying Eq. 11 to Eq. 8 and recalling the definition of $C_{pairs}$ in Eq. 7, we obtain

$$E[C] = \frac{1}{6}\sum_{u=1}^{n}\sum_{v=1}^{n} a_{uv} C_{pairs}(u,v) = \frac{C_{pairs}}{3}. \qquad (12)$$

The combination of Eq. 11 and Eq. 10 yields

$$E[C] = \frac{C_{pairs}}{3}. \qquad (13)$$

A simple case is a linear tree with $n = 4$, as $C_{pairs} = 1$ transforms Eq. 13 into $E[C]=1/3$.
Applying Eq. 3 to Eq. 13, one finally obtains

$$E[C] = \frac{n}{6}\left(n-1-\langle k^2 \rangle\right) \qquad (14)$$

for $n \geq 4$.
For the dependency tree of Fig. 1, $n = 9$ and $\langle k^2 \rangle = 4$ gives $E[C] = 6$.
According to Eq. 14, $E[C] = 0$ for a star tree as $\langle k^2 \rangle = n-1$ for that tree while

$$E[C] = \frac{n(n-5)}{6} + 1 \qquad (15)$$

for a linear tree as $\langle k^2 \rangle = 4 - 6/n$ in that case (Ferrer-i-Cancho 2013).

## 4. DISCUSSION

It has been shown that E[$C$] is determined exclusively by $n$ and $\langle k^2 \rangle$ (Eq. 14). Given $n$, the range of variation of E[$C$] is then given by $\langle k^2 \rangle$, which is minimum for a linear tree and maximum for a star tree, i.e. (Ferrer-i-Cancho 2013)

$$4 - \frac{6}{n} \leq \langle k^2 \rangle \leq n - 1 \qquad (16)$$

for a finite tree with $n \geq 2$ and thus giving

$$0 \leq E[C] \leq \frac{n(n-5)}{6} + 1 \qquad (17)$$

thanks to Eq. 15 for any tree of at least four vertices (E[$C$] = 0 if $n < 4$).

Fig. 4 shows the upper bound of E[$C$] provided by a linear tree (Eq. 17), which obviously grows asymptotically as $n^2$ for sufficiently large $n$. Thus the possibility that the rather small number of crossings of real sentences (Liu 2010) is the outcome of some sort of optimization processes, possibly a side-effect of the minimization of dependency lengths (Ferrer-i-Cancho 2006, Ferrer-i-Cancho 2013) cannot be denied. Future research on the significance of the small amount of crossings of real sentences should consider the real value of $C$ in sentences versus estimates of E[$C$] obtained through Eq. 14 with real values of $\langle k^2 \rangle$. Thus, investigating the scaling of $\langle k^2 \rangle$ as a function of $n$ in real sentences from dependency treebanks (e.g., Civit *et al.* 2006, Böhmová *et al.* 2003, Bosco *et al.* 2000) is an important question for future research.

The results presented above can also help to shed light on the actual relationship between dependency length and crossings (Ferrer-i-Cancho 2006, 2013, Liu 2008). Imagine that $\langle d \rangle$ is the mean dependency length of the linear arrangement of vertices. The possibility of a natural correlation between $C$ and $\langle d \rangle$ can be demonstrated starting from an actual sentence such as the one in Fig. 1 and swapping the position of pairs of vertices chosen at random. Fig. 5 shows that both $C$ and $\langle d \rangle$ start from $\langle d \rangle = 11/8 = 1.375$ and $C = 0$ for the sentence in Fig. 1 and then both increase as the number of these swaps increases till they converge to their values in a random linear arrangement, respectively, E[$C$] = 6 (computed above) and E[$\langle d \rangle$] = E[$d$] = $(n+1)/3$ = 10/3 ≈ 3.33 (Ferrer-i-Cancho 2004, 2013, Zörnig 1984). Notice that our swapping of vertex positions is a randomization procedure that preserves the dependency tree (i.e. the adjacency matrix of the tree), and thus preserves the degree's 2[nd] moment and the connectedness of the dependency network. Other research on dependency networks has employed procedures to generate random dependency structures that do not warrant that vertex degrees or connectedness are maintained (as needed by a tree) or forbid dependency crossings (Liu & Hu 2008).

Fig. 5 suggests that $C$ and $\langle d \rangle$ are positively correlated, which is consistent with the hypothesis that the low frequency of dependency crossings could be a side effect of dependency length minimization (Ferrer-i-Cancho 2006). Future research could extend this kind of analysis to more sentences with the help of dependency treebanks (e.g., Civit *et al.* 2006, Böhmová *et al.* 2003, Bosco *et al.* 2000).

Final note: the mathematical results presented in this article have been applied in a series of articles: Ferrer-i-Cancho (2014), Ferrer-i-Cancho (2016a,b), Esteban *et al.* (2016) and Gómez-Rodríguez & Ferrer-i-Cancho (2016).

**ACKNOWLEDGEMENTS**

We are grateful to O. Jiménez for helpful discussions and G. Wimmer, J. Baixeries, H. Liu and R. Cech for helpful comments on this manuscript. This work was supported by the grant *Iniciació i reincorporació a la recerca* from the Universitat Politècnica de Catalunya, the grants BASMATI (TIN2011-27479-C04-03) and OpenMT-2 (TIN2009-14675-C03) from MICINN (Ministerio de Ciencia e Innovación), the grant APCOM (TIN2014-57226-P) from MINECO (Ministerio de Economía y Competitividad) and the grant 2014SGR 890 (MACDA) from AGAUR (Generalitat de Catalunya).

## REFERENCES

**Böhmová, A., Hajič, J., Hajičová, E. & Hladká, B.** (2003). The Prague dependency treebank: three-level annotation scenario. In: Abeille, A. (ed.), *Treebanks: building and using syntactically annotated corpora.* Dordrecht: Kluwer, p. 103-127.

**Bollobás, B.** (1998). *Modern graph theory*. New York: Springer-Verlag.

**Bosco, C., Lombardo, V., Vassallo, D. & Lesmo. L.** (2000). Building a treebank for Italian: a data-driven annotation schema. In: *Proceedings of the 2nd International Conference on Language Resources and Evaluation LREC 2000, Athens, p. 99-105*.

**Civit, M., Martí, M.A. & Bufí, N.** (2006). 'Cat3LB and Cast3LB: from Constituents to dependencies'. In: *Advances in Natural Language Processing/* (LNAI, 4139), pp. 141-153. Berlin: Springer Verlag.

**Esteban, J.L., Ferrer-i-Cancho, R., & Gómez-Rodríguez, C.** (2016). The scaling of the minimum sum of edge lengths in uniformly random trees. *Journal of Statistical Mechanics, 063401*.

**Ferrer-i-Cancho, R.** (2004). Euclidean distance between syntactically linked words. *Physical Review E 70, 056135*.

**Ferrer-i-Cancho, R.** (2006). Why do syntactic links not cross? *Europhysics Letters 76, 1228-1235*.

**Ferrer-i-Cancho, R.** (2013). Hubiness, length, crossings and their relationships in syntactic dependency trees. *Glottometrics 25, 1-21*.

**Ferrer-i-Cancho, R.** (2014). A stronger null hypothesis for crossing dependencies. *Europhysics Letters 108 (5), 58003*.

**Ferrer-i-Cancho, R.** (2016a). Non-crossing dependencies: least effort, not grammar. In: Mehler, A., Lücking, A., Banisch, S., Blanchard, P. & Job, B. (eds.). *Towards a theoretical framework for analyzing complex linguistic networks.* Berlin: Springer, pp. 203-234.

**Ferrer-i-Cancho, R. & Gómez-Rodríguez, C.** (2016). Crossings as a side effect of dependency lengths. *Complexity 21 (S2), 320-328*.

**Gómez-Rodríguez, C. & Ferrer-i-Cancho, R.** (2016). The scarcity of crossing dependencies: a direct outcome of a specific constraint? http://arxiv.org/abs/1601.03210.

**Havelka, J.** (2007). Beyond projectivity: multilingual evaluation of constraints and measures on non-projective structures. In: *Proceedings of the 45th Annual Meeting of the Association of Computational Linguistics (ACL-07)*. Prague, Czech Republic: Association for Computational Linguistics, pp. 608-615.

**Hays, G.** (1964) Dependency theory: a formalism and some observations. *Language 40, 511-525.*

**Holan, T., Kubon, V., Plátek, M. & Oliva, K.** (2000). On complexity of word order. *Traitement automatique des langues 41 (1), 273-300.*

**Liu, H.** (2008). Dependency distance as a metric of language comprehension difficulty. *Journal of Cognitive Science 9, 159-191.*

**Liu, H.** (2010). Dependency direction as a means of word-order typology a method based on dependency treebanks. *Lingua. 120 (6), 1567-1578.*

**Liu, H. & Hu, F.** (2008). What role does syntax play in a language network? *Europhysics Letters 83, 18002.*

**Hudson, R.** (2000) Discontinuity. *Traitement automatique des langues 41 (1), 15-56.*

**Hudson, R.** (2007). *Language networks. The new word grammar.* Oxford University Press.

**Mel'čuk, I.** (1988). *Dependency syntax: theory and practice.* Albany, N.Y.: SUNY Press.

**Noy, M.** (1998). Enumeration of noncrossing trees on a circle. *Discrete Mathematics 180, 301-313.*

**Zörnig, P.** (1984). The distribution of the distance between like elements in a sequence. *Glottometrika 6, 1-15.*

# APPENDIX A: THE NUMBER OF POSSIBLE CROSSINGS OF AND EDGE

$C_{pairs}(u,v)$ can be derived from $C(u,v)$ assuming that $C(u,v;s,t)=1$ in any circumstance, which transforms Eq. 6 into

$$C(u,v) = C_{pairs}(u,v) = \frac{1}{2}\sum_{\substack{s=1\\s\neq u,v}}^{n}\sum_{\substack{t=1\\t\neq u,v}}^{n} a_{st} = \frac{1}{2}\sum_{\substack{s=1\\s\neq u,v}}^{n}(k_s - a_{us} - a_{vs}). \quad (A1)$$

Applying

$$\sum_{\substack{s=1\\s\neq u,v}}^{n} k_s = 2(n-1) - k_u - k_v \quad (A2)$$

and

$$\sum_{\substack{t=1\\t\neq u,v}}^{n} a_{st} = k_s - a_{us} - a_{vs} \quad (A3)$$

to Eq. A1 yields

$$C_{pairs}(u,v) = \frac{1}{2}(2(n-1) - k_u - k_v - (k_u - a_{uv} - a_{uu}) - (k_v - a_{uv} - a_{vv})) \quad (A4)$$

and finally

$$C_{pairs}(u,v) = (n-1) - k_u - k_v + a_{uv} + a_{uu} + a_{vv} = n - k_u - k_v \quad (A5)$$

as $a_{uu} = a_{vv} = 0$ (loops are not allowed) and $a_{uv} = 1$ as $u$ and $v$ are linked by the definition of $C(u,v)$.

# APPENDIX B: LINEAR AND STAR TREES HAVE UNIQUE DEGREE 2$^{nd}$ MOMENT

To simplify the arguments below, we define the degree 2$^{nd}$ moment as $\langle k^2 \rangle = K_2/n$, where $K_2(n)$, is the sum of squared degrees of a tree of $n$ vertices, i.e.

$$K_2(n) = \sum_{i=1}^{n} k_i^2. \quad (B1)$$

$K_2^{linear}(n)$ and $K_2^{star}(n)$ are defined, respectively, as the sum of squared degrees of a linear tree and a star tree of $n$ nodes. $K_2^{linear}(n) = 4n - 6$ and $K_2^{star}(n) = n(n-1)$ (Ferrer-i-Cancho 2013). Here it will be shown that a linear tree is the only tree for which $K_2(n)$ reaches $K_2^{linear}(n)$ while a star tree is the only tree for which $K_2(n)$ can reach $K_2^{star}(n)$. Before proving these properties, the concept of tree reduction and compact definitions of star and linear trees will be introduced.

*Tree reduction*

Any tree of at least two vertices has at least two leaves (Bollobás 1998, p. 11). Thus, any tree of $n + 1$ vertices ($n \geq 2$ is assumed) can be reduced to a tree of $n$ vertices by removing one of its leaves. Notice that this reduction will never disconnect the tree as the leaf removed cannot be attached to another leaf unless $n = 2$ (a leaf attached to another leaf when $n>2$ would contradict that a tree is a connected graph). Consider that the leaf removed is attached to a vertex of degree $k$ in the original tree (the tree of $n + 1$ vertices). Then

$$K_2(n + 1) = K_2(n) + k^2 - (k - 1)^2 + 1 \qquad (B2)$$

for the original tree and thus

$$K_2(n + 1) = K_2(n) + 2k. \qquad (B3)$$

*A star tree is a tree with a vertex of maximum degree.*

A star tree of $n$ vertices is a tree with a vertex of degree $n$-1 and $n$-1 leaves (Fig. 3). Indeed, a star tree of $n$ vertices can simply be defined as a tree with a vertex of maximum degree (i.e. degree $n$-1). The point is that the fact that a vertex has degree $n$-1 implies that there are $n$-1 leaves. To see it, recall that the degree sequence of a graph of $n$ vertices satisfies

$$\sum_{i=1}^{n} k_i = 2(n-1). \qquad (B4)$$

Assuming without any loss of generality that the *n-th* vertex has maximum degree (i.e. $k_n = n-1$), Eq. B4 gives

$$\sum_{i=1}^{n-1} k_i = n - 1 \qquad (B5)$$

As a tree is a connected graph, any vertex has degree greater than zero and Eq. B5 gives $k_1=...=k_i=...=k_{n-1}=1$, i.e. $n$-1 leaves, the number of leaves of a star tree.

*A linear tree is a tree where all vertex degrees do not exceed two*

A linear tree is a tree where all vertices have degree two except two leaves (Fig. 3). Indeed, a linear tree can simply be defined as a tree where all vertex degrees do not exceed two. Notice that in our last definition of linear tree we do not need to state the number of leaves and the number of vertices of degree two that we have. To understand our last definition of linear tree, suppose that a tree has $n$ vertices and $m$ leaves (then it has $m - 2$ vertices of degree 2). Then the sum of the degrees of leaves is $m$. If no vertex degree exceeds two then the sum of degrees of the vertices that are not leaves is $2(n - m)$. Then, Eq. B4 reduces to $m + 2(n - m) = 2(n - 1)$ which gives $m = 2$. Thus, if no vertex degree exceeds two, one can be certain that the tree is linear.

*A star tree is the only tree reaching $K_2^{star}(n)$*

Next it will be shown that a star tree is the only tree for which $K_2(n) = K_2^{star}(n)$. If $n = 2$, then this is trivially true as the only possible tree is a star tree. Consider a tree of $n+1$ vertices (with $n > 2$) such that $K_2(n + 1) = K_2^{star}(n + 1)$. Thanks to Eq. B3, we know that

$$K_2(n) + 2k = K_2^{star}(n + 1) \tag{B6}$$

for that tree. Adding that $K_2(n) \leq K_2^{star}(n)$ (Ferrer-i-Cancho 2013) to Eq. B6, it is obtained

$$k \geq \frac{K_2^{star}(n+1) - K_2^{star}(n)}{2} = \frac{(n+1)n - n(n-1)}{2} = n. \tag{B7}$$

As $k$ cannot exceed $n$ in a graph of $n + 1$ vertices (without loops), Eq. B7 implies that $k = n$, which we have shown above to imply that the tree of $n + 1$ vertices is a star, as we wanted to prove.

*A linear tree is the only tree reaching $K_2^{linear}(n)$*

Next it will be shown that a linear tree is the only tree for which $K_2(n) = K_2^{linear}(n)$. If $n = 2$, then this is trivially true as the only possible tree is a linear tree. Consider a tree of $n+1$ vertices (with $n > 2$) such that $K_2(n + 1) = K^{linear}(n + 1)$. Thanks to Eq. B3, we know that

$$K_2(n) + 2k = K_2^{linear}(n + 1) \tag{B8}$$

for that tree. Adding that $K_2(n) \geq K_2^{linear}(n)$ (Ferrer-i-Cancho, 2013) to Eq. B8, it is obtained

$$k \leq \frac{K_2^{linear}(n+1) - K_2^{linear}(n)}{2} = \frac{4(n+1) - 6 - (4n-6)}{2} = 2. \tag{B9}$$

As $k$ is the degree of a vertex that is not a leaf, if follows that any vertex in the original tree that is not a leaf has degree exactly 2, which we have shown above to imply that the tree of $n+1$ vertices is a linear tree, as we wanted to prove.

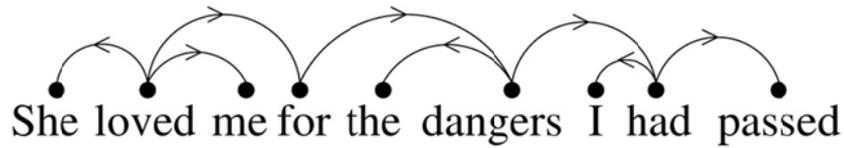

**Figure 1.** The syntactic structure of the sentence *'She loved me for the dangers I had passed'* following the conventions in (Mel'čuk 1988). *'she'* and the verb *'loved'* are linked by a syntactic dependency. Arcs go from governors to dependents. Thus, *'she'* and *'me'* are dependents of the verbal form *'loved'*. Indeed, *'she'* and *'me'* are arguments of the verb form *'loved'* (the former as subject and the latter as object).

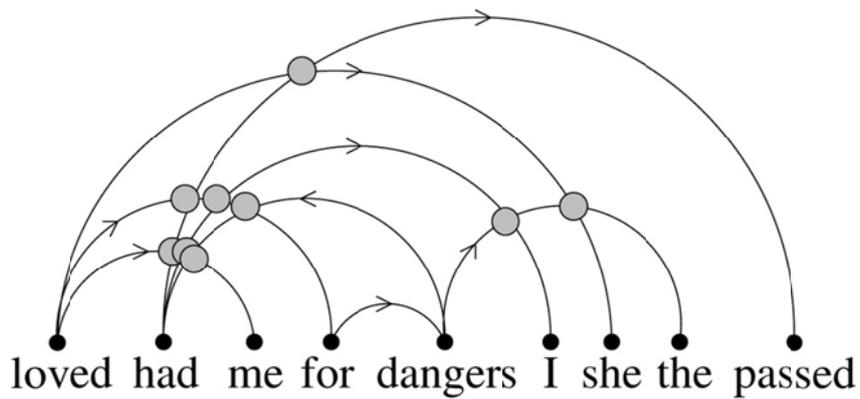

**Figure 2.** The structure of the sentence in Fig. 1 after a random linear rearrangement of its words. Gray circles indicate edge crossings.

(a) (b)

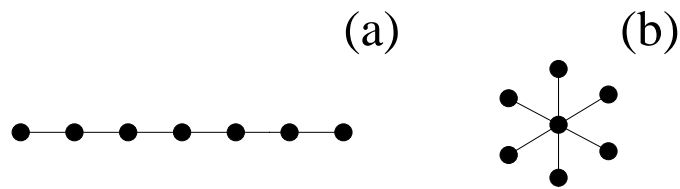

**Figure 3.** (a) a linear tree and (b) a star tree. A linear tree is a tree with the smallest possible number of leaves (only two leaves, Bollobás 1998, p. 11) while a star tree is the tree with the largest number of leaves.

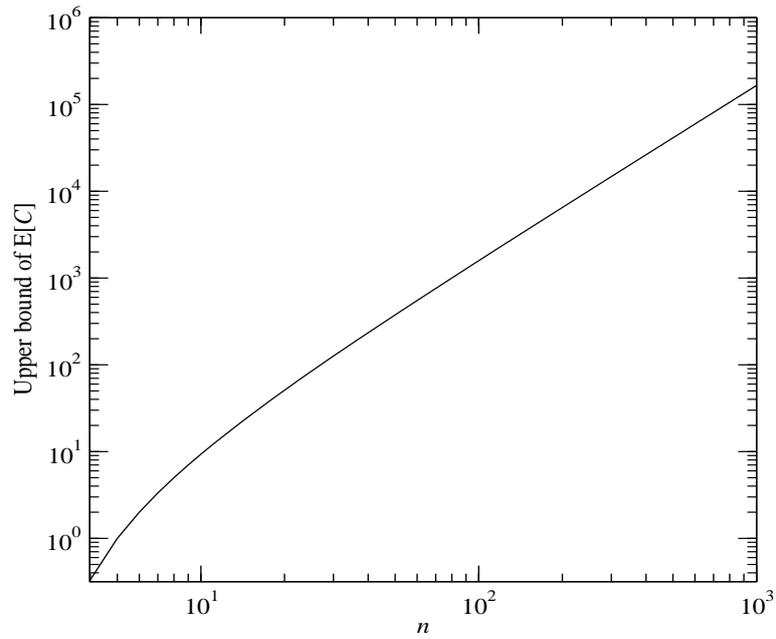

**Figure 4.** The upper bound of E[*C*] (the expectation of the number of crossings of a linear tree) as function of *n*, the number of vertices of the tree.

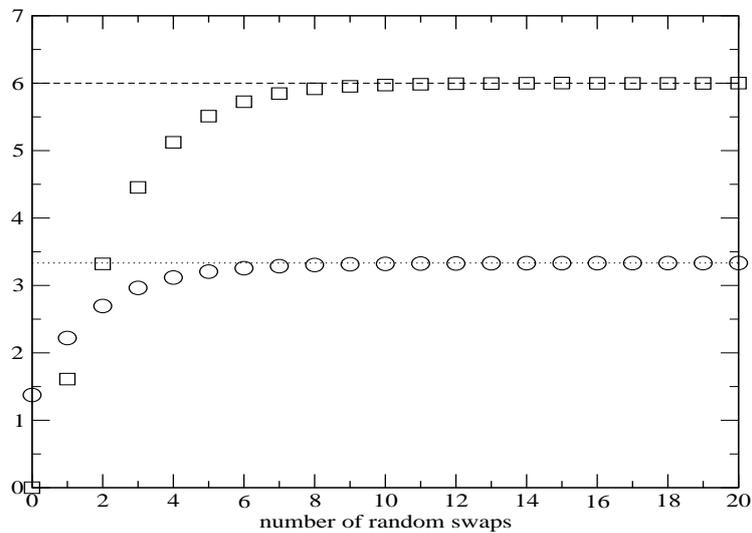

**Figure 5.** The evolution of $\langle d \rangle$, the mean dependency distance (circles), and *C*, the number of edge crossings (squares), versus the number of swaps of pairs of vertex positions for the sentence in Fig. 1. Each curve is the average over $10^6$ replicas. $\langle d \rangle$ converges to $E[d] = 10/3$ (dotted line) while *C* converges to $E[C] = 6$ (dashed line).